\documentclass[11pt,a4paper]{article}
\usepackage{emnlp2018}
\usepackage{times}
\usepackage{latexsym}
\usepackage{amsmath}
\usepackage{amssymb}
\usepackage{graphicx}

\usepackage{color}
\usepackage{multirow}

\usepackage{url}
\usepackage{footnote}
\usepackage{footmisc}

\aclfinalcopy % Uncomment this line for the final submission
\def\aclpaperid{171} %  Enter the acl Paper ID here

\title{Mapping Text to Knowledge Graph Entities\\using Multi-Sense LSTMs}

\author{Dimitri Kartsaklis\footnotemark{} \quad\quad\quad Mohammad Taher Pilehvar\quad\quad\quad Nigel Collier \vspace{0.2cm}\\
Department of Theoretical and Applied Linguistics\\
University of Cambridge\\
Cambridge, UK\\
{\tt \{dk426;mp792;nhc30\}@cam.ac.uk}}

\date{}

\newcommand{\astfootnotetext}[1]{
\let\oldthefootnote=\thefootnote
\setcounter{footnote}{0}
\renewcommand{\thefootnote}{\fnsymbol{footnote}}
\footnotetext{#1}
\let\thefootnote=\oldthefootnote
}

\begin{document}

\maketitle

\begin{abstract}
This paper addresses the problem of mapping natural language text to knowledge base entities. The mapping process is approached as a composition of a phrase or a sentence into a point in a multi-dimensional entity space obtained from a knowledge graph. The compositional model is an LSTM equipped with a dynamic disambiguation mechanism on the input word embeddings (a {\em Multi-Sense LSTM}), addressing polysemy issues. Further, the knowledge base space is prepared by collecting random walks from a graph enhanced with textual features, which act as a set of semantic bridges between text and knowledge base entities. The ideas of this work are demonstrated on large-scale text-to-entity mapping and entity classification tasks, with state of the art results.
\end{abstract}

\urlstyle{same}

\astfootnotetext{\hspace{-0.110 cm}$^*$ This paper is dedicated to the memory of Euripides Kartsaklis, a man who loved technology.}

\section{Introduction}

The task of associating a well-defined action, concept or piece of knowledge to a natural language utterance or text is a common problem in natural language processing and generic artificial intelligence \cite{tellex2011}, and can emerge in many different forms. In NLP, the ability to code text into an entity of a knowledge graph finds applications in tasks such as question answering and information retrieval, or any task that involves some form of mapping a definition to a term \cite{hill,relpron}. Further, it can be invaluable in providing solutions to domain-specific challenges, for example medical concept normalisation \cite{nut2016} and identification of adverse drug reactions \cite{oconnor2014}.

This paper details a model for efficiently mapping unrestricted text at the level of phrases and sentences to the entities of a knowledge base (KB)---a task also referred to as text {\em grounding} or {\em normalisation}. The model aims at characterising short focused texts, such as definitions or tweets. Given a medical KB, for example, a tweet of the form ``Can't sleep, too tired to think straight'' would be mapped to the entity {\em Insomnia}, while in the context of a lexical ontology the definition ``Device that detects planets'' would be associated to the entity {\em Telescope}. 

Note that such a task cannot be approached as standard classification, since the ``classes'' (entities) are usually in one-to-one correspondence with the available inputs. To address this we propose the use of a continuous vector space for embedding the entities of the KB graph, where text is projected by a neural network. We rely on the notion of distributional semantics, where a word is represented as a multi-dimensional vector obtained either by collecting co-occurrence statistics with a selected set of contexts or by directly optimising an objective function in a neural network-based architecture \cite{collobert2008}. Interestingly, similar techniques can be used for the multi-dimensional representation of nodes in a KB graph; for example, by collecting random walks following the edges of a graph it is possible for one to construct an artificial ``corpus'', to which a distributional model applies in the usual way \cite{deepwalk}.

By exploiting this representational compatibility, we treat the process of text-to-entity mapping as a transformation from a textual vector space where words live, to a KB vector space created from a graph and populated by vectors representing entities. A sentence is coded as a sequence of word vectors, composed by a modified Long Short-Term Memory  network (LSTM---\cite{lstm}) into a multi-dimensional point in the entity space. One of our aims is to specifically deal with lexical ambiguity and polysemy which can be an important factor for the task at hand. To this end, each word is associated with a number of sense embeddings, and the LSTM is extended with an attentional disambiguation mechanism that dynamically selects and updates the right sense vector for each word given its context during training. We dub this formulation {\em Multi-Sense LSTM} (MS-LSTM).

% To the best of our knowledge, this is the first time that such an approach is used for the purpose of text-to-entity mapping.\footnote{Of course, vector space transformation approaches have been exploited before in multilingual settings.} Especially, the use of a quantitative representation of KB entities brings a number of advantages, the most important of which is that its continuous nature allows for smooth decision boundaries in a variety of tasks, something generally not possible with more traditional methods, e.g. string matching. 

% Furthermore, there is a ready-made arsenal of NLP techniques that can be used for training the vectors, while the plethora of available online resources guarantees sufficient coverage of training data. 

An important issue is the provision of a set of reliable anchors; that is, points in one-to-one correspondence between the two representations that would enforce some degree of structural similarity between pieces of text and KB entities and thus make the mapping more efficient. We deal with this problem by extending the original KB graph with nodes corresponding to {\em textual features}, i.e. to words strongly associated to a specific entity and collected from various resources. A novel sampling strategy is detailed for incorporating these nodes to random walks, which are then fed to the skipgram model for producing an entity space. The results indicate that the textual nodes, being words {\em and} KB entities at the same time, do an extremely effective job in transforming the geometry of the entity space to the benefit of mapping the textual modality. 

The proposed model is evaluated in three tasks: text-to-entity mapping on a dataset extracted from {\sc Snomed CT}\footnote{\url{https://www.snomed.org/snomed-ct}}, a medical knowledge base of 327,000 concepts; a reverse dictionary task based on WordNet \cite{wordnet}, where the goal is to associate a multi-word definition to the correct lemma \cite{hill}; and document classification on the Cora dataset \cite{cora}. The results  demonstrate the effectiveness of our methods by improving the current state of the art.

%We close this section by summarising our contributions: (1) We propose an approach to semantic grounding of arbitrary text to ontological knowledge based on multi-dimensional representations of concepts; (2) we detail a methodology for generating this KB space by a graph extended with textual features attached to normal concept nodes; (3) we present a compositional model supported by a dynamic word sense disambiguation mechanism able to transform a piece of textual information to a point into the KB space.

\section{Background}
\label{sec:relwork}

Aligning meaning between text and entities in a knowledge graph is a task traditionally based on heuristic methods exploiting text features such as string matching, word weighting, syntactic relations, or dictionary lookups  \cite{mccallum2005,lu2011,oconnor2014}. Machine learning techniques have been also exploited in various forms, for example \newcite{leaman} use a pairwise learning-to-rank technique to learn the similarity between different terms, while \newcite{nut2015} apply statistical machine translation to ``translate'' social media text to domain-specific terminology. There is little work based on neural networks; the most relevant to us is a study by \newcite{hill}, who tested a number of compositional neural architectures trained to approximate word embeddings on a reverse dictionary task. Compared to their work, this paper proposes the use of a distinct target space for representing ontological knowledge, where every entity in the graph lives.

The goal of a graph embedding method is to embed components of a knowledge graph into a low-dimensional space. One research direction focuses on the relations, i.e. the edges of the graph \cite{bordes2013,socher2013,transg} and aims at tasks such as link prediction and KB completion. Such work is outside the scope of the current paper, the subject of which is the efficient low-dimensional representation of {\em entities} (nodes). In this line of research, a prevalent method involves the collection of a set of random walks, starting from each node in the graph \cite{deepwalk,node2vec}. There is a direct analogy between such a set of random walks and a text corpus: each node corresponds to a word and the sequence of nodes visited during a random walk is  analogous to a sentence. Thus, any distributional model that takes as input this artificial ``corpus'' can generate multi-dimensional representations of the nodes in the graph. Random walks have also been used for KB inference \cite{lao2011} with success. 

While random walk-based methods are not the only way to construct graph spaces---alternatives include factorisation \cite{ahmed} and deep autoencoders \cite{wang2016}---they have been found very effective in capturing multiple aspects of the graph structure \cite{wang2017,goyal2017}. The current paper proposes a random walk generation strategy that improves and complements existing approaches.

The idea of using textual features to improve the entity vectors is not well explored, and most of the existing work focuses again on the representation of relations \cite{xie2016,wang2014,wang-tf} as opposed to entities. Closer to us is the work of \newcite{yamada} and \newcite{yang2015}, with the latter to incorporate text features in the concept embeddings by exploiting matrix factorisation properties. 

% In this paper we follow the more intuitive approach to treat textual features as first-class citizens of the graph, that is, nodes attached to related concepts with weighted edges. In this way we preserve the generality of our methods without introducing extra optimisation overhead.

Representing the meaning of words using a number of sense vectors is an old and well-established idea in NLP---see for example \cite{Schutze,reisinger2010,neelakantan}. However, most of the relevant research is evaluated on intrinsic tasks such as word similarity, while in the few works based in real end tasks, disambiguation is usually treated as a prior stand-alone step \cite{kartsaklis:2013:EMNLP,li2015,pilehvar2017}. The crucial difference of this work is that the ambiguity resolution mechanism is part of the compositional model itself, and the sense embeddings are trained simultaneously with the rest of the parameters. A close work is by \newcite{cheng2015}, who used a siamese network with an integrated disambiguation mechanism for paraphrase detection. For more information on multi-sense embeddings see \cite{camacho2018}.

\section{Methodology}
\label{sec:overview}

\begin{figure}
\centering
\includegraphics[trim={0 0.5 0.5 0.35cm},clip,scale=0.26]{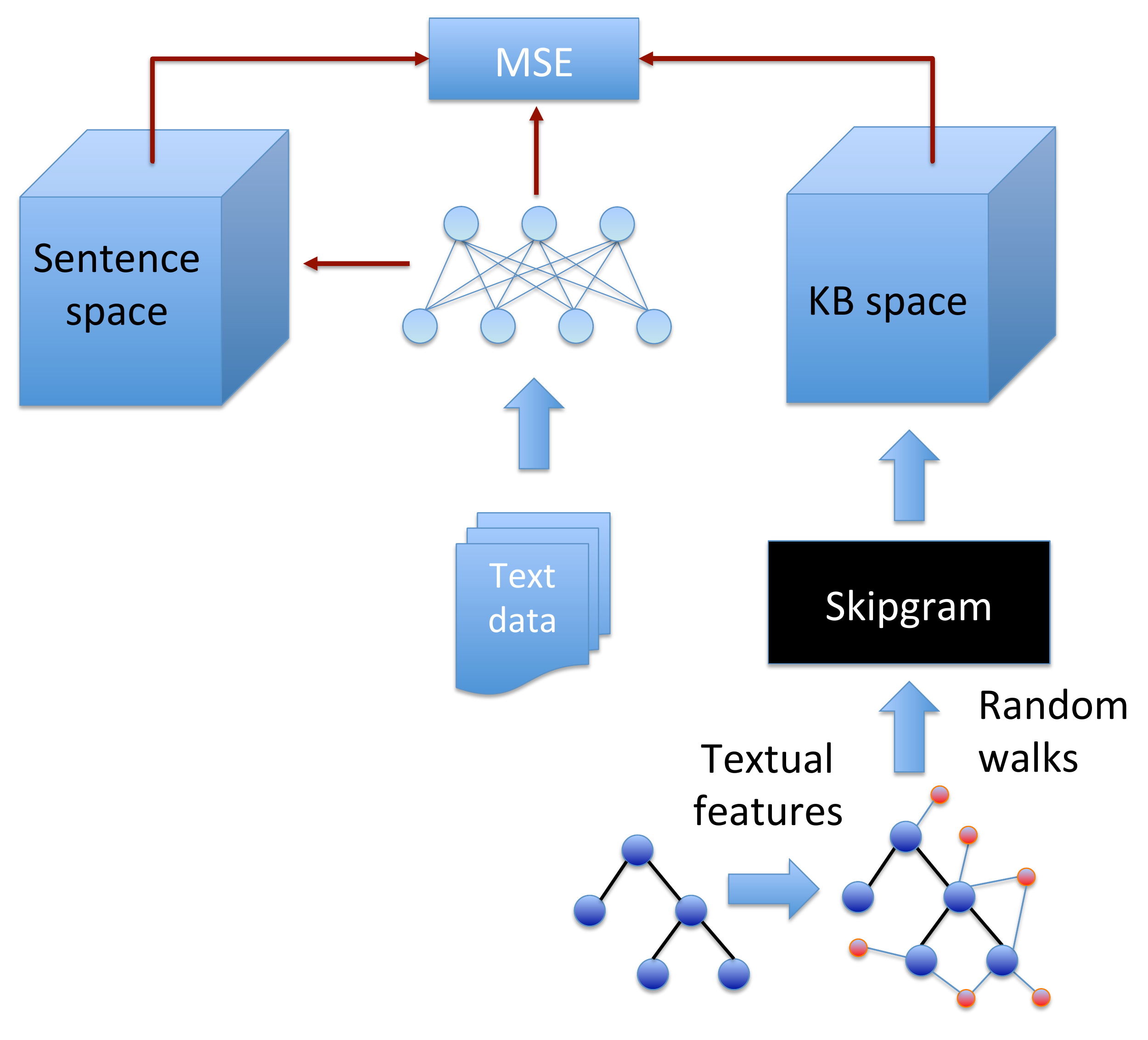}
\caption{The text-to-entity mapping system in a nutshell. The red nodes indicate textual features, while ``MSE'' stands for mean squared error.}
\label{fig:generic}
\end{figure}

Fig. \ref{fig:generic} provides a high-level illustration of our methodology, consisting of two stages: (1) the KB graph is extended with weighted textual features, and an artificial ``corpus'' of random walks is created and used as input to the skipgram model \cite{mikolov2013efficient} for generating an enhanced KB space---this part is covered in \S \ref{sec:text}; (2) the transformation from text to entities is performed by a supervised multi-sense compositional model, which generates a point in the KB space for every input text. This is achieved with an LSTM recurrent network, equipped with an attentional mechanism that provides a finer level of granularity to the different ways a word is used in the data---we detail this part in \S \ref{sec:model}.

%\begin{enumerate}
%  \item\label{enu:step1} The KB graph is extended with weighted textual features, and an artificial ``corpus'' of random walks is created; this corpus is used as input to the skipgram model \cite{mikolov2013efficient} for generating an enhanced KB space.
%  \item\label{enu:step2} The transformation from text to concepts is achieved via a supervised multi-sense compositional model, which generates a point in the KB space for every input text.
%\end{enumerate}

% The first stage deals with the fact that natural language text and ontological knowledge reflect different aspects of reality that need to be brought closer, and is discussed in detail in \S \ref{sec:text}. The compositional model of the second stage is a {\em long short-term memory} (LSTM) recurrent network \cite{lstm} acting on learnable word embeddings and trained to approximate the vector of the correct concept in the concept space. The model is further equipped with an attentional mechanism (not shown in Figure \ref{fig:generic}) that provides a finer level of granularity to the different ways a word is used in the data (\S \ref{sec:model}).

\subsection{Textual features for entity vectors}
\label{sec:text}

For our KB space, we follow the generic recipe proposed by \newcite{deepwalk} and we assemble an artificial corpus of random walks from the KB graph, which is then used as input to the skipgram model \cite{mikolov2013efficient}. For a random walk of nodes $n_1n_2\dots n_T$ and a context window size $c$, skipgram maximises the following quantity:

\vspace{-0.2cm}
%\small
\begin{equation}
\frac{1}{T}\sum\limits_{t=1}^T \sum\limits_{-c\leq j \leq c, j\neq 0} \log p(n_{t+j}|n_t)
\end{equation}
\normalsize

\noindent i.e. for a target node $n_t$, the objective is to predict all other nodes in the same context. As a consequence, two vectors of the resulting space will be close if their corresponding nodes occur in topological proximity within the graph. However, while such a topology allows perhaps for meaningful comparisons between points in this space, it is not directly compatible with the task of mapping text to entities. The reason is that the communities formed in a KB graph (and thus the topology of the resulting vector space) mostly reflect domain-specific hierarchies and ontological relationships that are not necessarily evident by the textual representations referring to the entities. An important question therefore with regard to the proposed methodology is how to provide meaningful links between the two representations that would allow for the efficient translation of one form (text) to another (entities).

%\begin{figure}      
%\includegraphics[trim={4cm 1.5cm 3cm 1.5cm},clip,scale=0.24]{eye.png}
%  \caption{Comparing the neighbours for word ``eye'' in a standard word space (left) with those of the SNOMED concept ``eye (body part)'' in a graph embedding space (right).}
%  \label{fig:topology}
%\end{figure}

In this work, we address this problem by associating every node in the graph with a set of textual features, each one of which is weighted according to their importance with respect to the node. Our methodology is as follows: For each entity, we collect all available textual descriptions found in the knowledge base itself and the English portion of BabelNet \cite{babelnet}, which is a very large dictionary integrating numerous resources, such as WordNet, Wikipedia, FrameNet and many others. The textual descriptions are treated as short documents, and each word in them is assigned a specific TF-IDF value, forming the set of textual features for the specific entity. 

The KB graph is extended in the following way: Let $T_c$ be the set of textual features for an entity $c$; then, for each $t$ in $T_c$, we add an edge $(c,t)$ with weight $\text{tf-idf}_c(t)$, where tf-idf$_c(t)$ is the TF-IDF value of $t$ with respect to $c$.
In contrast to \newcite{deepwalk} who utilise a uniform node sampling strategy, we define the random walk generation process as follows: Given a randomly selected node $n$, let $C_n = \{c_1,c_2,\cdots c_N\}$ be the set of all entity nodes in its immediate vicinity, and $T_n = \{t_1,t_2,\cdots t_M\}$ the set of all textual features of $n$; the next node $x$ in the path is drawn from a categorical distribution defined as below:

\vspace{-0.3cm}
\small
\begin{equation}
  P_X(x) = \left\{
     \begin{array}{lr}
       (1-\lambda) \dfrac{1}{N} & \text{if}~x \in C_n \vspace{0.2cm}\\
       \lambda \dfrac{\text{tf-idf}_{n}(t_i)}{\sum_{j=1}^{M} \text{tf-idf}_{n}(t_j)} &
       \text{if}~x \in T_n \\
     \end{array}\right.
\normalsize
\label{equ:prob}
\end{equation}
\normalsize

%\vspace{-0.4cm}
%\small
%\begin{equation}
%p(c_i) = \frac{1-\lambda}{N} ~~~~~~
%p(t_i) = \lambda \frac{\text{tf-idf}_{n_t}(t_i)}{\sum_{j=1}^{M} \text{tf-idf}_{n_t}(t_j)} 
%\label{equ:prob}
%\end{equation}
%\normalsize

\noindent for $X$ a discrete random variable with range $C_n \cup T_n$. In the above, $\lambda$ defines the proportion of the probability mass allocated to textual features, when both $C_n$ and $T_n$ are non-empty; if one of the sets is empty, all of the probability mass is allocated to the other set, and $\lambda$ becomes irrelevant. Further, in contrast to what is the case for the textual nodes, the probabilities of the entity nodes in Equation \ref{equ:prob} are defined uniformly, since we lack any mechanism for fine-tuning them in a way that objectively reflects the importance of the nodes.

\begin{figure}      
\includegraphics[trim={0 0.5cm 0 0.5cm},clip,scale=0.4]{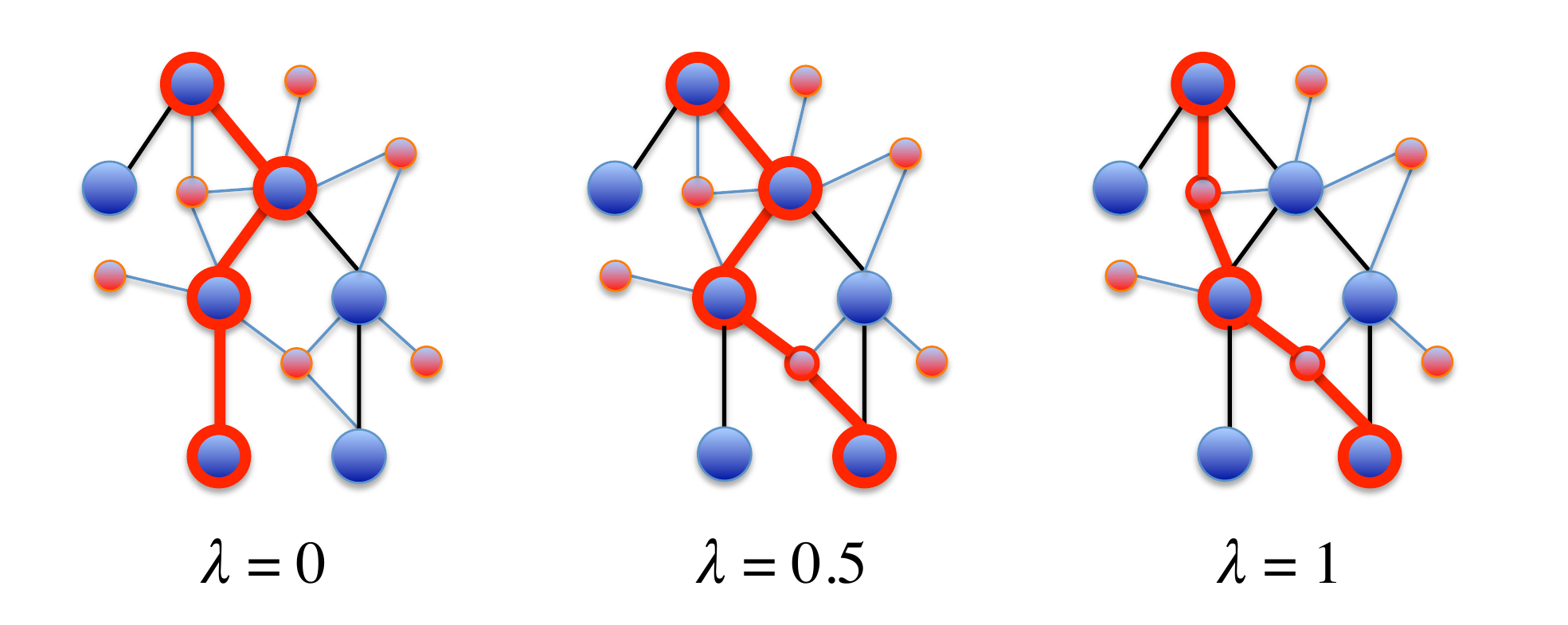}
  \caption{Effect of $\lambda$ parameter. Blue nodes indicate entities, red nodes textual features, and red paths refer to random walks. As $\lambda$ increases, the probability of ``hops'' between originally unlinked nodes increases accordingly.}
  \label{fig:sampling}
\end{figure}

It is instructive to examine how the above sampling strategy works. As expected, setting $\lambda=0$ will result in a sampling process that ignores the textual features and produces a path comprised solely of entity nodes; this is equivalent to the original model by \newcite{deepwalk}, known as DeepWalk. On the other hand, the effect of setting $\lambda=1$ is less intuitive: Recall that, by construction, each textual node is connected only to entity nodes; that is, when the current node is textual, the next node will be always sampled from $C_n$. Therefore, setting $\lambda=1$ creates  paths following an alternating pattern, where each entity node is followed by a textual node, which in turn is followed by an entity node. Values of $\lambda$ between 0 and 1 scale this behaviour accordingly (Fig. \ref{fig:sampling}).

% Equation \ref{equ:prob} defines a weighting scheme that can be readily used in conjunction with other node sampling methods. \newcite{node2vec}, for example, introduce in the random walk generation process a search bias $\alpha$ that provides a trade-off between breadth-first and depth-first searches; the unnormalized probability of selecting a node $x$ from a current node $v$ is given as $\alpha \cdot w_{vx}$, where $w_{vx}$ is the weight of the $(v,x)$ edge, and which in our case is computed as in Equation \ref{equ:prob}.

% and one could indeed argue that certain relationships in the graph (e.g. of ``is-a'' type) should be weighted differently than others (e.g. of ``has-a'' type). Although this is reasonable, since we lack any mechanism to fine-tune the weights of the concept nodes in a way that objectively reflects their importance, we do not further pursue the issue in this paper.

\begin{figure}
  \centering
  \includegraphics[trim={0 0 0 0},clip,scale=0.30]{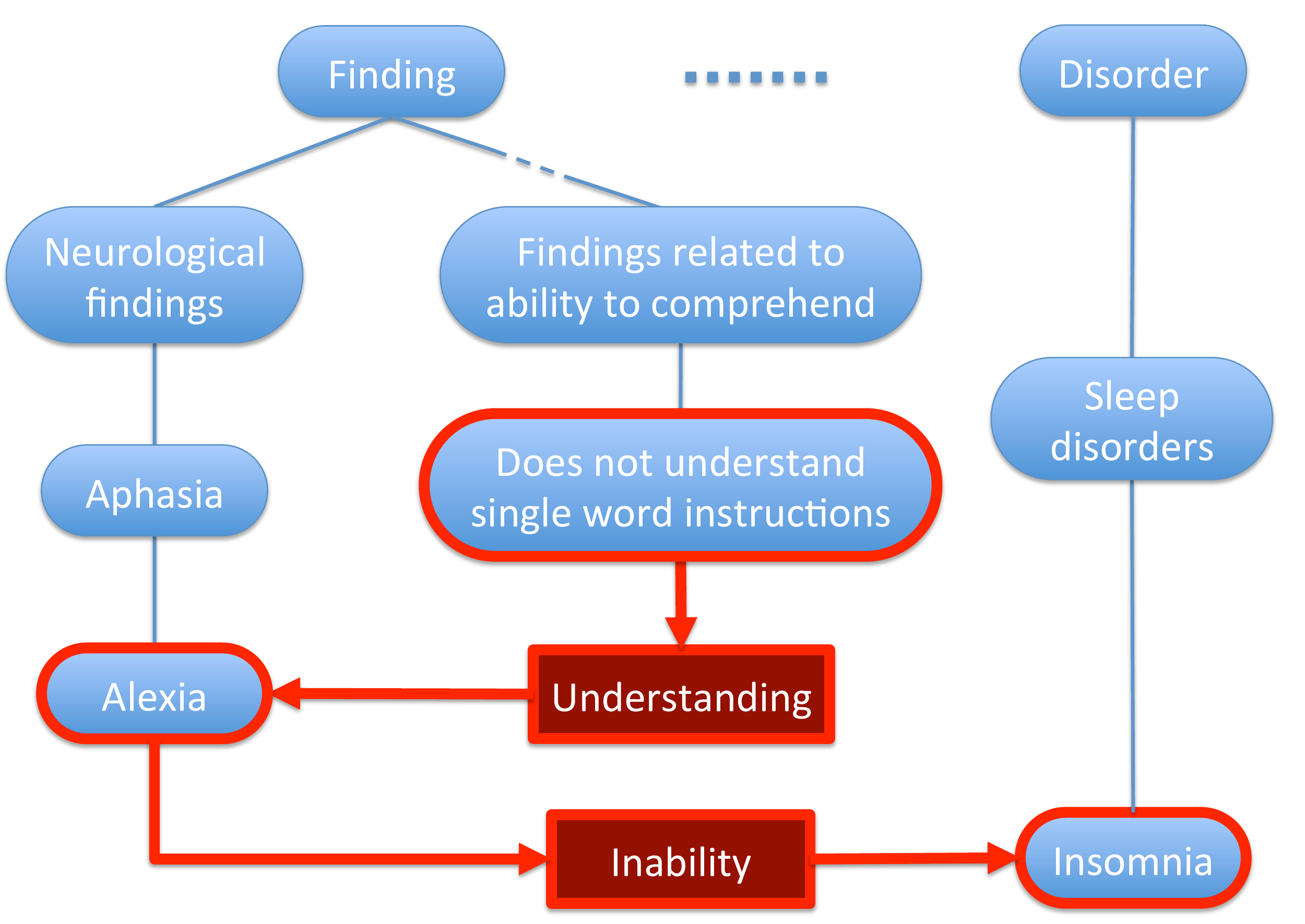}
  \caption{Linking of distant concepts with textual features (red boxes) for $\lambda=1$. The textual feature {\em understanding} correctly links a related medical finding (lying on a different branch) to the condition known as {\em alexia}. Further, due to the presence of {\em inability} in their contexts, the vectors of {\em alexia} and {\em insomnia} (concepts originally quite far apart in the graph) will now have a common part reflecting that they are both conditions related to forms of incompetence.}
  \label{fig:example}
\end{figure}

\paragraph{Advantages.} The introduction of textual features in the graph achieves two goals. Firstly, the textual nodes serve as links between entities which, although perhaps related to each other in some way, lie in different parts of the KB graph (e.g. being parts of different hierarchies). As a result, points that would normally be unjustifiably apart of each other in the vector space are now brought closer, providing additional coherence. This behaviour is controlled by the $\lambda$ parameter, as Fig. \ref{fig:sampling} shows. Fig. \ref{fig:example} presents an illustrative example, taken from a real random walk on {\sc Snomed CT}. 

The second advantage of introducing textual features in the graph is a consequence of the dual nature of these features in the context of learning: they essentially represent words, but since they are also nodes of the graph, they get vector representations exactly as every other normal entity in the knowledge base. The textual features, therefore, paired with their assigned vectors, form a set of {\em anchors} that links pieces of text with the KB space, and can be used to support the training process of the mapping system. In \S \ref{sec:grounding} we will see that this approach leads to substantial improvements in the accuracy of the model.

\subsection{A multi-sense LSTM}
\label{sec:model}

We now proceed to present our neural architecture for text-to-entity mapping. The goal of the model is, given a certain piece of text, to produce a point in the KB space corresponding to an appropriate entity or concept. The model is trained on pairs of texts and entity vectors created from a graph extended with textual features, as discussed in \S \ref{sec:text}.  

Our architecture needs to explicitly take into account the fact that the task at hand is very sensitive to lexical ambiguity. Specifically, while it is true that the level of homonymy (words having more than one disjoint meanings) is substantially decreased when we move from the generic domain to more specialised domains, on the other hand the increase in polysemy (words with many slightly different meanings) is exponential. As an example, while the lemma for the word ``fever'' in a dictionary usually contains two or three definitions, the term occurs in many dozens of different forms and contexts in {\sc Snomed}. Note that most of the different uses of the term correspond to distinct KB nodes, a fact that makes the job of a text-to-entity mapping system especially hard.\footnote{See also \S\ref{sec:qualitative} for some concrete examples.}
This motivates the employment of a dedicated mechanism that would handle the extra complexity imposed by the polysemous words. 

\begin{figure}[t]
  \centering
  \includegraphics[trim={0.8cm 0.8 1.5cm 1.5},clip,scale=0.255]{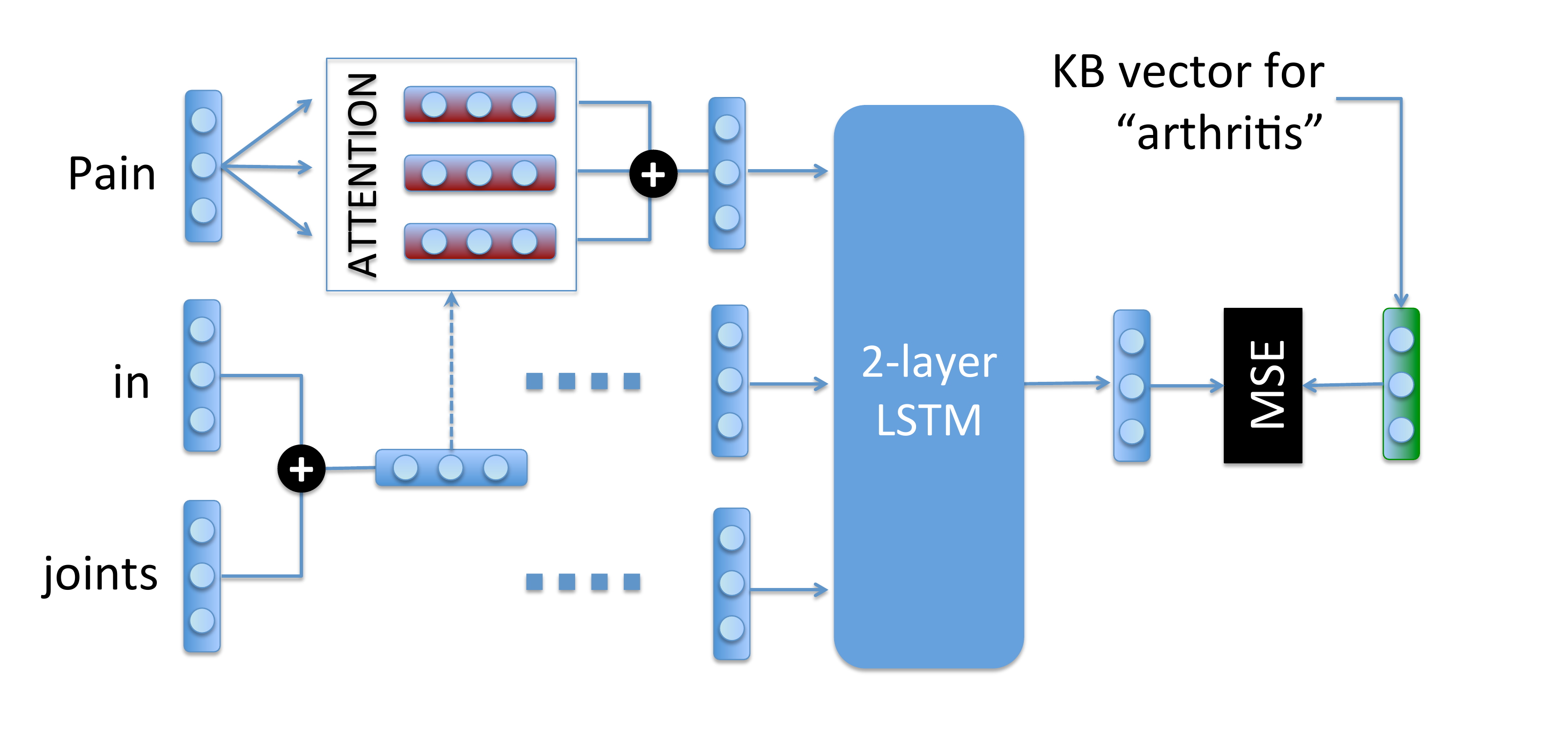}
  \caption{Detailed architecture of the Multi-Sense LSTM (MS-LSTM), shown here for the first word of a phrase. Red vectors refer to senses, while the green vector is a target vector. 
  %In the example, each sense of word {\em pain} will be associated with a different kind of medical condition based on the available contexts.
  }
  \label{fig:model}
\end{figure}

The compositional setting of this paper, equip\-ped with such a mechanism, is shown in Fig. \ref{fig:model}. It consists of a generic word embedding layer, a word sense disambiguation layer, and two consecutive LSTM networks responsible for encoding the embeddings into a vector in the KB space.  The objective is to minimise the mean squared error between the predicted vectors and the target vectors (prepared as in \S \ref{sec:text}):

\vspace{-0.2cm}
%\small
\begin{equation}
   \text{MSE} = \frac{1}{N}\sum\limits_{i=1}^N \| y_i - f(x_i) \|^2
   \label{equ:mse}
\end{equation}
\normalsize

\noindent where $N$ is the number of training examples, $x$ the input text, $y$ the target entity vector, and $f$ the neural network.

To address the polysemy issues discussed above, every word is associated with a single generic embedding and $k$ sense embeddings, where $k$ is a fixed number. These sense embeddings can be seen as centroids of clusters denoting different uses of the word in the training set, and are dynamically updated during training. Specifically, for each word $w_i$ in a training example, a context vector $c_i$ is computed as the average of the {\em generic} vectors of all other words in the sentence. The probability of each sense vector $s_{ij}$ given this context is then calculated via an attentional mechanism equipped with a softmax layer, as follows: 

%\small
\vspace{-0.2cm}
\begin{equation}
  p(s_{ij}|c_i) = \frac{\exp{( {w'_j}^\intercal s'_{ij})}}{\sum_{l=1}^k \exp{( {w'_l}^\intercal s'_{il})}}
\end{equation}
\normalsize

\noindent where $s'_{ij} = \tanh(W s_{ij} + U c_i)$, and $W$, $U$ and $W'$ the parameters of the attentional network. Each sense vector is subsequently updated by addition of the context vector weighted by its similarity with the specific sense:

%\small
\vspace{-0.2cm}
\begin{equation}
  s^{t+1}_{ij} = s^t_{ij}+ ( {s^t_{ij}}^\intercal c^t_i) c^t_i 
\end{equation}
\normalsize

The output of the attention is a weighted sum of the sense vectors given their probabilities (i.e. we apply {\em soft attention}), which is used as input to the compositional network---a 2-layer LSTM. The overall model is optimised on the MSE of the LSTM's output vector and the target entity vector. At inference time, a predicted vector $\hat{y}$ can be classified to the entity with the closest vectorial representation according to some metric.

\section{Experiments}
\label{sec:exp}

The ideas presented in the previous sections are evaluated on three tasks, two of which are related to text-to-entity mapping, and one to classification of KB entities. The purpose of the classification task (\S\ref{sec:cora}) is to provide a direct comparison of the textually enhanced vectors against vectors produced by the original graph, but {\em independently} of the compositional part. On the other hand, the text mapping experiments (\S\S \ref{sec:grounding}, \ref{sec:revdict}) evaluate the overall architecture of Fig. \ref{fig:generic} (including the compositional model and the dynamic disambiguation mechanism) on appropriate end tasks. Comparisons are provided with the most relevant previous work. Specifically, in all tasks, no inclusion of textual features corresponds to the standard DeepWalk model of \newcite{deepwalk}; in \S\ref{sec:revdict} our compositional architecture is compared to the work of \newcite{hill} in their reverse dictionary task; and \S\ref{sec:cora} compares our method for textually enhancing the entity space with that of \newcite{yang2015}, and other state-of-the-art deep models. The last subsection, \S\ref{sec:qualitative}, examines a few selected cases from a qualitative perspective.

\subsection{Text-to-entity mapping}
\label{sec:grounding}

We begin with a large scale text-to-entity mapping experiment. We construct a dataset of 21,000 medical concepts extracted from {\sc Snomed CT}, each of which is associated with a multi-word textual description, taken from the knowledge base or BabelNet. The criterion for including a concept in the dataset was the availability of at least one textual description with 4 or more words. The objective of the task is to associate each one of these descriptions to the correct concept. Given a predicted vector $\hat{v}$, we assemble a list of all candidate concept vectors ranked by their cosine similarity with $\hat{v}$. We compute strict accuracy (based on how  many times the vector of the correct concept is at the top of the list) and accuracy on the first 20 elements of the list. Further, we also present results based on the mean reciprocal rank (MRR).

%\vspace{-0.1cm}
%\small
%\begin{equation}
%  \text{MGD} = \frac{1}{N} \sum\limits_{i=1}^N \frac{1}{\text{shortest-path}(\hat{c_i},c_i)}
%\end{equation}
%\normalsize
%
%In the above, $N$ is the size of the testing dataset, and the shortest-path function returns 1 if $\hat{c}=c$, or the number of nodes in the shortest path between them (with the two nodes inclusive) otherwise. Finally, we also present results based on the mean reciprocal rank (MRR).
%, a metric used in information retrieval for evaluating the quality of responses to web queries.
%
%\begin{equation}
%  \text{MRR} =  \frac{1}{N} \sum\limits_{i=1}^N \frac{1}{\text{rank}(\hat{c})}
%\end{equation}
%

In all experiments, we create KB vectors of 150 dimensions by applying the skipgram objective on a set of random walks of length 20, and with window size of 5. The graph is extended with 102,500 textual nodes weighted by their TF-IDF values with regard to the corresponding entities and selected as described in \S \ref{sec:text} (textual features that occur in the testing set are not taken into account). Each node in the graph serves as the starting point of 10 random walks. For the compositional model, we use embeddings of 150 dimensions, and 200-dimensional hidden states. The attentional mechanism is implemented as a 2-layer MLP, with 50 units allocated to the hidden layer for each sense. The overall model contains two dropout layers for regularisation purposes, and is optimised with Adam \cite{adam} ($\alpha=0.001$, $\beta_1=0.9$, $\beta_2=0.999$).\footnote{Python code will be released at \url{https://github.com/cambridgeltl/SIPHS}.}

Following usual practice, we split our dataset in three parts: a training set (14,754 instances), a testing set (4,187 instances), and a development set (2,000 instances). We use the dev set to optimise the two main hyper-parameters of our model, namely the probability mass given to textual features ($\lambda$) and the number of senses for each word ($k$). The experiments on the dev set showed that increasing the probability mass for the inclusion of textual features in the random walks leads to consistently better performance for all tested models, so for the main experiment we set $\lambda$ to its highest possible value, $1.00$.\footnote{Recall that this means half of the nodes in a random walk will be textual (see Fig. \ref{fig:sampling}).} Further, a number of senses equal to 3 achieved the highest performance. 

We compare our MS-LSTM with a number of baselines: In {\em Baselines 1} and {\em 2} a vector for each textual description is computed as the average of pre-computed word vectors, and compared to concept vectors prepared in a similar way, i.e. by averaging pre-computed vectors for all words in the qualified name of the entities. We used two different word spaces, a standard Word2Vec space created from Google News\footnote{\url{https://code.google.com/archive/p/word2vec}} and a custom Word2Vec model trained on a corpus of 4B tokens from medical articles indexed in PubMed\footnote{\url{http://www.ncbi.nlm.nih.gov/pubmed/}}. In {\em Least squares} and {\em CCA}, an averaged vector for each textual description is again computed as before, and a linear mapping is learned between the textual space and the KB space, using least squares and canonical correlation analysis \cite{cca}. 

In {\em Standard LSTM}, we use a configuration similar to that of Fig. \ref{fig:model}, but {\em without} the multi-sense aspect; here, the word embeddings are just parameters of the model randomly initialised before training. Further, we also test a standard LSTM where the length of the single embeddings is $k$ times bigger ($k$ is the number of senses in the MS-LSTM), so that the overall dimensionality of embeddings in LSTM and MS-LSTM is the same.

\begin{table}
\footnotesize
\centering
\setlength{\tabcolsep}{2.7pt}
\begin{tabular}{|c|c|ccc|}
\hline
  {\bf Model} & {\bf Target space} & MRR & Acc & Acc-20 \\
\hline\hline      
Baseline 1 & W2V-GoogleNews & 0.25 & 0.19 & 0.41 \\
Baseline 2 & W2V-PubMed     & 0.17 & 0.12 & 0.31 \\
\hline
\multirow{2}{*}{Least squares} & DeepWalk & 0.19 & 0.10 & 0.49 \\
 & TF vectors & {\bf 0.49} & {\bf 0.37} & {\bf 0.79} \\
\hline
\multirow{2}{*}{CCA} & DeepWalk & 0.36 & 0.24 & 0.70   \\
  & TF vectors & {\bf 0.71} & {\bf 0.60} & {\bf 0.94} \\
\hline
Standard LSTM & DeepWalk & 0.30 & 0.20 & 0.58 \\
 (150 dim.) & TF vectors & {\bf 0.82} & {\bf 0.73} & {\bf 0.97} \\
\hline
Standard LSTM & DeepWalk & 0.33 & 0.23 & 0.59 \\
 ($k\times 150$ dim.) & TF vectors & {\bf 0.86} & {\bf 0.80} & {\bf 0.97} \\
\hline
\multirow{2}{*}{MS-LSTM} & DeepWalk & 0.36 & 0.26 & 0.60 \\
  & TF vectors & {\bf 0.89} & {\bf 0.84} & {\bf 0.98} \\
\hline
MS-LSTM & \multirow{2}{*}{TF vectors} & \multirow{2}{*}{{\bf\underline{0.94}}} & \multirow{2}{*}{{\bf\underline{0.90}}} & \multirow{2}{*}{{\bf\underline{1.00}}} \\
+ anchors & & & & \\
   
\hline
\end{tabular}
\normalsize

\caption{Results for the {\sc Snomed} dataset. For the MS-LSTM we set $k=3$, while {\em TF vectors} refers to our textually enhanced vectors ($\lambda=1$). The difference between MS-LSTM and LSTM is s.s. with $p<0.01$ according to a two-tailed $z$-test.}
\label{tbl:exp1}

\end{table}

The results are presented in Table \ref{tbl:exp1}. Each model is tested against two target KB spaces, one consisting of simple DeepWalk vectors\footnote{In our setting, this is equivalent to having $\lambda=0$.} and one of textually enhanced vectors (TF vectors, $\lambda=1$) according to the procedure of \S\ref{sec:text}. There are three observations: (1) Using the enhanced vectors as a target space improves the  performance of {\em all} tested models by a large margin; (2) the MS-LSTM configuration of Fig. \ref{fig:model} achieves the highest overall performance, showing that explicitly handling polysemy during the composition is beneficial for the task at hand; and (3) despite the equal dimensionality between the two models, the standard LSTM with the long embeddings presents performance inferior to that of the MS-LSTM.

The last row of the table presents results after extending the training dataset with the textual anchors, that is, all the textual features paired with their learned KB vectors, as described in the {\em Advantages} section in \S\ref{sec:text}. Specifically, recall that each textual feature (a word or a two-word compound), being also a node in the graph, is associated with a vector according to the process of \S\ref{sec:text}. It is possible for one then to use these ({\em textual feature}, {\em vector}) pairs as additional examples during the training of the MS-LSTM. The last row of Table \ref{tbl:exp1} shows the results after extending the training set with the 102,500 textual features. This setting achieves the highest performance, increasing further the strict accuracy by 6\%, to 0.90. 

\subsection{Reverse dictionary}
\label{sec:revdict}
%\vspace{-0.1cm}

We proceed to the reverse dictionary task of \newcite{hill}, the goal of which is to return a candidate term given a definition. Many forms of this task have been proposed in the past, see for  example \cite{kartsaklis2012,turney2014,relpron}. In \cite{hill}, the authors test a number of supervised models under two evaluation modes: (1) ``seen'', in which the testing instances are also included in the training set; and (2) ``unseen'', where the evaluation is done on a held-out set. In both cases the datasets consisted of 500 term-definition pairs from WordNet. 

We treat WordNet as a graph, the edges of which are defined by the various relationships between the synsets. This graph is further extended with 96,734 textual nodes extracted from the synset descriptions. We compute synset vectors of 150 dimensions, on random walks of length 20 and with window size of 5.  For the seen evaluation, we train the compositional model on the totality of WordNet 3.0 synsets (117,659) and their descriptions. For the unseen evaluation, we remove from the graph any textual features occurring in the testing part, and create a new set of synset vectors; further, any testing instance is removed from the training set of the compositional model. The evaluation is done by comparing the predicted vector with the vectors of all WordNet synsets (a search space of 117,659 points) and creating a ranked list as before, by cosine similarity. Following \cite{hill}, we compute accuracy on top-10 and top-100. $\lambda$ and $k$ are tuned on a dev set of 2,000 synsets, showing a behaviour very similar to that of the {\sc Snomed} task.

\begin{table}
\footnotesize
\setlength{\tabcolsep}{4pt}
\centering
\begin{tabular}{|c|cc|}
\hline
 {\bf Model} & {\bf Acc-10} & {\bf Acc-100} \\
\hline\hline
\multicolumn{3}{|c|}{\bf Seen (500 WordNet definitions)} \\
\hline\hline
OneLook \cite{hill}  & 0.89 & 0.91 \\
RNN cosine \cite{hill} & 0.48 & 0.73 \\
\hline
% MS-LSTM, $\lambda=0.00$, $k=1$ & 0.05 & 0.24 & 0.50 \\
% MS-LSTM, $\lambda=0.50$, $k=1$ & 0.50 & 0.77 & 0.92 \\
% Standard LSTM, $\lambda=1.00$ & 0.64 & 0.86 & 0.96 \\ 
Std LSTM (150 dim.) + TF vec. & 0.86 & 0.96 \\ 
Std LSTM ($k\times 150$ dim.) + TF vec. & 0.93 & 0.98 \\
%\hline    
%MS-LSTM, $\lambda=1.00$, $k=3$ & {\bf 0.76} & {\bf 0.92} & {\bf 0.97} \\
\hline
MS-LSTM +TF vectors & {\bf 0.95} & {\bf 0.99} \\
MS-LSTM +TF vectors + anchors & {\underline {\bf 0.96}} & {\underline {\bf 0.99}} \\
\hline\hline
\multicolumn{3}{|c|}{\bf Unseen (500 WordNet definitions)} \\
\hline\hline
RNN w2v cosine \cite{hill} & 0.44 & 0.69 \\
BOW w2v cosine \cite{hill} & 0.46 & 0.71 \\
\hline
% Standard LSTM, $\lambda=1.00$ & 0.48 & 0.72 & 0.88 \\
Std LSTM (150 dim.) + TF vec. & 0.72 & 0.88 \\
Std LSTM ($k\times 150$ dim.) + TF vec. & 0.77 & 0.90 \\
% MS-LSTM, $\lambda=1.00$, $k=3$ & {\bf 0.50} & {\bf 0.77} & {\bf 0.92} \\
\hline
MS-LSTM + TF vectors & {\bf 0.79} & {\bf 0.90} \\
MS-LSTM + TF vectors + anchors & {\underline {\bf 0.80}} & {\underline {\bf 0.91}}  \\
\hline       
\end{tabular}
\caption{Results for the reverse dictionary task, compared with the highest numbers reported by \newcite{hill}. {\em TF vectors} refers to textually enhanced vectors with $\lambda=1$. For the MS-LSTM, $k$ is set to 3.}
\label{tbl:wordnet}
\end{table}

Table \ref{tbl:wordnet} shows the results, based on a MS-LSTM setup similar to that of \S\ref{sec:grounding}. Note that the MS-LSTM achieves 0.95-0.96 top-10 accuracy for the seen evaluation, significantly higher not only than the best model of \newcite{hill}, but also higher than OneLook, a commercial system with access to more than 1000 dictionaries. It also presents considerably higher performance in the unseen evaluation. We are not aware of any other models with higher performance on the specific task.

\subsection{Document classification}
\label{sec:cora}

Our last experiment is a document classification task, performed on Cora \cite{cora}, a dataset containing 2708 machine learning papers linked by citation relationships into a graph. Each document is a short text extracted from the title or the abstract of the paper. The task is to predict the category of a document (a total of 7 classes), given its vector---so here we only evaluate the textually enhanced vectors as inputs to a classifier, independently of the compositional part. 

In Table \ref{tbl:cora} we report results for two evaluation settings. In Evaluation 1, we provide a comparison with the method of \newcite{yang2015} who include textual features in graph embeddings based on matrix factorisation, and two topic models used as baselines in their paper. Using the same classification algorithm (a linear SVM) and training ratio (0.50) with them, we present state-of-the-art results for vectors of 150 dimensions, prepared by a graph extended with 1422 textual features. We set $\lambda=0.5$ by tuning on a dev set of 677 randomly selected entries from the training data.\footnote{We also attempted a second classification experiment on a dataset of 200k concepts extracted from {\sc Snomed}, observing a similar behaviour of $\lambda$ (details are not reported due to space). This difference in the behaviour of $\lambda$ between text-to-entity mapping and classification tasks is discussed in \S\ref{sec:discussion}. }

\begin{table}
\footnotesize
\centering
\setlength{\tabcolsep}{15pt}
\begin{tabular}{|c|c|}
\hline
{\bf Model} & {\bf Accuracy}  \\
\hline\hline
\multicolumn{2}{|c|}{\bf Evaluation 1 (training ratio=0.50)} \\
\hline\hline
PLSA \cite{hofmann1999} & 0.68 \\
NetPLSA \cite{mei2008} & 0.85 \\
TADW \cite{yang2015} & 0.87 \\
\hline
Linear SVM + DeepWalk vectors & 0.85 \\
%Linear SVM+TF vectors, $\lambda=0.25$ & 0.880 \\
Linear SVM + TF vectors & {\bf 0.88} \\
%Linear SVM+TF vectors, $\lambda=0.75$ & 0.844 \\
%Linear SVM+TF vectors, $\lambda=1.00$ & 0.750 \\

\hline\hline
\multicolumn{2}{|c|}{\bf Evaluation 2 (training ratio=0.05)} \\
\hline\hline

Planetoid \cite{yang2016revisiting} & 0.76 \\
GCN \cite{kipf2016semi} & 0.81 \\
GAT \cite{gan} &  {\bf 0.83} \\
\hline
Linear SVM + DeepWalk vectors & 0.72 \\
Linear SVM + TF vectors & {\bf 0.82} \\
\hline
\end{tabular}

\caption{Results for the Cora dataset. {\em TF vectors} refers to textually enhanced KB vectors ($\lambda=0.5$). Difference between our best models and GAT/GCN/TADW are {\em not} s.s.}
\label{tbl:cora}
\normalsize
\end{table}

\begin{table*}
  \footnotesize
  \setlength{\tabcolsep}{8pt}
  \centering
  
  \begin{tabular}{|c|cc|}
  \hline
  {\bf Definition from the unseen dataset of the reverse dictionary task} & $k=3$ (correct pred.) & $k=1$ (wrong pred.) \\
  \hline
  the branch of engineering that deals with things smaller than 100 nm & nanotechnology & microelectronics \\
  floor consisting of open space at the top of a house just below roof & loft & balcony \\
   a board game for two players; pieces move according to dice throws & backgammon & checkers \\
  an address of a religious nature & sermon & rogation \\
%  a small town in southern Pennsylvania & Gettysburg & Altoona (city in central Penns.) \\
  \hline
  \end{tabular}
  
  \setlength{\tabcolsep}{7.1pt}
  \begin{tabular}{|c|cc|}
  \hline
  {\bf Example short phrase with ambiguous words} & $k=3$ prediction & $k=1$ prediction \\
  \hline
  a rechargeable cell & nickel-cadmium battery & karyolysis ({\em biological process}) \\
  a state capital & Curitiba ({\em Brazilian state capital}) & assert ({\em verb}) \\
  the lap of a person & upper side of thighs & lapper ({\em garment}) \\
  a band named Queen & band leader & neckband ({\em garment}) \\
  \hline
  \end{tabular}
  \caption{Qualitative comparison of a few selected cases for multi- and single-sense LSTMs.}
  \label{tbl:qualitative}
\end{table*}

In Evaluation 2, using the same linear SVM classifier and $\lambda$ as before, we reduce the training ratio to 0.05 in order to make our task comparable to the experiments reported by \newcite{gan} for a number of  deep learning models: specifically, the graph attention network (GAT) of \newcite{gan}, the graph convolutional network (GCN) of \newcite{kipf2016semi}, and the Planetoid model of \newcite{yang2016revisiting}. Again, our simple setting presents results within the state of the art range, comparable to (or better than) those of much more sophisticated models that have been specifically designed for the task of node classification. We consider this as a strong indication for the effectiveness of the textually enhanced vectors as representations of KB entities.

\begin{figure}
  ~\includegraphics[scale=0.35, trim=3.64cm 1.74cm 15cm 1.9cm, clip]{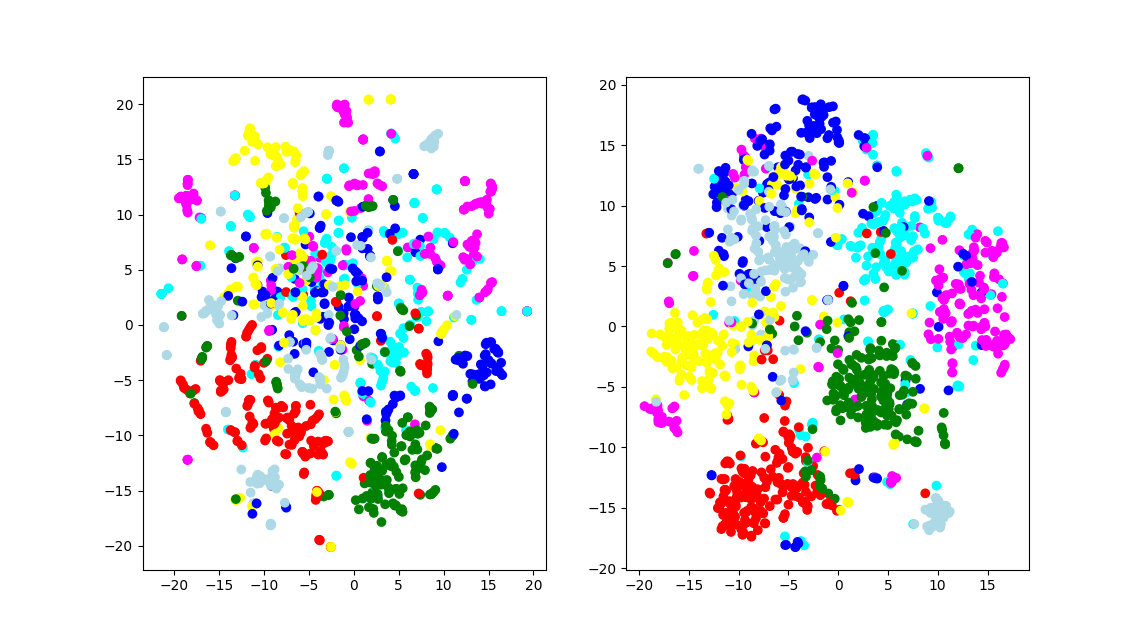}~~
  \includegraphics[scale=0.35, trim=15.9cm 1.74cm 2cm 1.9cm, clip]{cora.png}
  \caption{Visualisation of the Cora classes based on a 2D t-SNE projection of the node vectors before the inclusion of textual features (left) and after (right).}
  \label{fig:cora}
\end{figure}

Fig. \ref{fig:cora} provides a visualisation of the Cora clas\-ses based on node vectors created with $\lambda=0$ and $\lambda=0.5$, correspondingly, demonstrating the impact of textual features in terms of cluster coherence and separation. 

%We tune $\lambda$ on a development set of 677 randomly selected entries; this time the highest performance comes from $\lambda=0.50$, while it is interesting to note that, for the specific task, too many textual features are harmful, compared to including no textual features at all; this finding is discussed in detail in \S\ref{sec:discussion}.\footnote{We also attempted another classification experiment on a dataset of 82,000 medical concepts extracted from SNO\-MED, observing a very similar behaviour of $\lambda$. The details are not reported here due to lack of space.} 

%providing a comparison with the method of \newcite{yang2015} who include textual features in graph embeddings based on matrix factorisation, and two topic models used as baselines in their paper. Using the same classification algorithm and training ratio with them, we present results for vectors of 150 dimensions, prepared by a graph extended with 1422 textual features. 

%We also compare our vectors to a number of recent works: specifically, the graph attention network (GAT) of \newcite{gan}, the graph convolutional network (GCN) of \newcite{kipf2016semi}, and the Planetoid model of \newcite{yang2016revisiting}. All of our configurations for $\lambda > 0$ improve the current state of the art for the dataset. This time the highest performance comes from $\lambda=0.50$, while it is interesting to note that, for the specific task, too many textual features are harmful, compared to including no textual features at all; the findings are discussed in detail in \S\ref{sec:discussion}.

\subsection{Qualitative evaluation}
\label{sec:qualitative}
% \vspace{-0.1cm}

Table \ref{tbl:qualitative} compares the performance of the multi-sense approach with that of the single-sense model for a number of selected cases of text mapping. The predictions in the top part (for definitions taken from the unseen evaluation of the reverse dictionary task) show that, in contrast to the single-sense model, the multi-sense approach was able to capture subtle variations of meaning between different synsets due to polysemy, as motivated in \S\ref{sec:model}. The lower part of the table contains short phrases with ambiguous words, specifically selected to demonstrate the effect of the multi-sense approach. In all these cases, the multi-sense model was able to effectively disambiguate the ambiguous parts of the phrase by using the available context, and predict a very relevant synset; in contrast, the predictions of the single-sense model were based on choosing a wrong sense. 

Finally, Table \ref{tbl:senses} presents the derived senses for word {\em table}, expressed as lists of nearest neighbouring words in the space. The model was able to effectively distinguish between a table as a kitchen furniture (sense 2), and a table as a structured way of presenting data (senses 1 and 3).

\section{Discussion}
\label{sec:discussion}

The experimental work shows that using a graph embedding space as a target for mapping text to entities is an effective approach. This was mostly evident in the reverse dictionary task of \S\ref{sec:revdict}, where the model was found to perform substantially better than previous approaches by \newcite{hill}, who used a compositional architecture similar to ours but optimised on the {\em word} embeddings of the target terms. Note that this is sub-optimal in the sense that, unless specific measures are taken, a word embedding reflects ambiguous meaning; therefore, trying to associate a definition like ``keyboard musical instrument with pipes'' to the vector for word ``organ'' introduces a certain amount of noise in the model, since the definition will be partly associated with features related to the ``body part'' sense of the word. In our model, homonymy issues are resolved by design: each point in the target space corresponds to a well-defined unambiguous concept or synset. Further, the attentional mechanism of Fig. \ref{fig:model} handles subtle variations of each distinct sense due to polysemy.

The effectiveness of the textual feature mechanism was demonstrated in every task we attempted, but to different extents. As our tuning on the dev sets showed, for tasks closer to text-to-entity mapping (\S\S \ref{sec:grounding}-\ref{sec:revdict}) the more the textual features in the random walks, the better the results were. However, the best performance on the classification task came by $\lambda$ values between 0.50 and 0.75, i.e. by walks visiting more entity nodes than textual nodes. The reason is that entity classification is a task very sensitive to the topology of the KB graph, since entities belonging to a specific class are very likely to be located at the same sub-hierarchy, hence in topological proximity. On the other hand, one of the motivations for introducing textual features was exactly to broaden the context of a node by connecting distant parts of the graph (see Figures \ref{fig:sampling}-\ref{fig:example}). So, while small amounts of textual features can be still useful for classification purposes, excessive use introduces unwanted noise in the model.

\begin{table}
\centering
\small
\begin{tabular}{p{7.2cm}}
\hline
{\bf Sense 1.} formulation, uncommonly, rauwolfia, cardiology, hypodermic, malleability, points, optic, dendrite, rubiaceae, nonparametric, meninges, deviation, anesthetics\\
%ingrowth modeling disable administration philosophies appositive surveyor's investigation acutely used marie bandage chiasm subsidy attestation skull sandglass polarities inclement scurvy fbi phases completion suspenseful discerned vulnerability jambs chemists aorist statesmanlike flaunted crippling \\
\hline
{\bf Sense 2.} tableware, meal, expectation, heartily, kitchen, hum, eating, forestay, suitors, croupier, companionship, restaurant, dishes, candles, cup, tea
\\
% demography valuables  smoldering sympathomimetic fondness leave affection furnished meals accessing tidy candles eat associations cup monovalent belonging cups enclose emit sips mechanic liqueurs unexplored floats tea sweeter moustache sadness pronouns sarawak inhabitants include arrange \\
\hline
{\bf Sense 3.} reassigned, projective, ultracentrifuge, polemoniaceous, thyronine, assumptions, lymphocyte, atomic, difficulties, intracellular, virgil, elementary, cartesian 
\\
%adds adenosine beset featherfoil centrifugation interchangeability palmyra smoldering dissonance inhibition volume hum markup pending atavist breathless osmosis hydrolyses uncertainty perturbed scalars oscillator modeling quantum risked estimating titration optically methodical saves marketable molecules auditing monovalent devise \\
\hline
\end{tabular}
\normalsize
\caption{Derived senses for word {\em table}, visualised as lists of nearest neighbouring words in the vector space.}
\label{tbl:senses}
\end{table}

The dynamic disambiguation mechanism integrated in the compositional architecture improved further the performance of the model. This finding is consistent with previous work on simpler tensor-based models, which showed that applying some form of word sense disambiguation when composing word vectors can provide consistent improvements on end tasks such as sentence similarity and paraphrase detection \cite{kartsaklis:2013:EMNLP,kartsaklis:2013:CoNLL}. 

% The dynamic disambiguation mechanism integrated in the compositional architecture improved further the performance of the model. While there are ways to lift the restriction of imposing a fixed number of senses for each word, we feel that this would unnecessarily complicate the training process, since in our model a variable number of sense embeddings can be simulated by setting $k$ to a reasonable high value (see also discussion in \S\ref{sec:grounding}). 

% in Table \ref{tbl:exp1}, for example, we can see that setting $k=10$ performs almost the same as setting $k=3$ (the best model in the task).

%, as the results in Table \ref{tbl:exp1} show for different $k$s. 

%\vspace{-0.1cm}
\section{Conclusion and future work}
\label{sec:conclusion}
%\vspace{-0.1cm}

%We presented and thoroughly evaluated a semantic grounding system based on a continuous KB space enhanced with textual features and capable of handling polysemy. As for future work, having a mechanism that translates arbitrary text to points in a continuous space creates many  opportunities for interesting research. For example, while  the size of a knowledge base is finite, the space itself consists of infinite number of points, each one of which corresponds to a valid, yet not explicitly stated in the KB, concept of the same domain. The exciting question of how can we exploit this extra information---for example in order to {\em enrich} the knowledge base with new data---constitutes one of our future directions. 

We presented and evaluated a text-to-entity mapping system based on a continuous KB space enhanced with textual features and capable of handling polysemy. The reasonable next step will be to extend our methods for modelling the {\em relations} (edges) of a KB graph, which will allow applications in tasks such as link prediction and KB completion. Furthermore, having a mechanism that translates arbitrary text to points in a continuous space creates many  opportunities for interesting research. For example, while  the size of a knowledge base is finite, the space itself consists of infinite number of points, each one of which corresponds to a valid---yet not explicitly stated in the KB---entity of the same domain. The exciting question of how can we exploit this extra information---for instance in order to {\em enrich} the knowledge base with new data---constitutes one of our future directions. 

\section*{Acknowledgments}

The authors would like to thank Victor Prokhorov and Ehsan Shareghi for useful discussions on the paper, as well as the anonymous reviewers for their suggestions. 
The idea of a dynamic disambiguation mechanism embedded in a neural architecture originally appeared in \cite{cheng2015} as part of a generic multi-prototype embedding model. Kartsaklis and Collier acknowledge support by EPSRC grant EP/M005089/1; Pilehvar and Collier are supported by MRC grant No. MR/M025160/1. We are also grateful to NVIDIA Corporation for  the donation of a Titan XP GPU.

%%%%%%%%%%
% References and End of Paper \bibliography{Bibliography-File}
% \newpage
\bibliography{refs.bib}
\bibliographystyle{acl_natbib}

\end{document}